\RequirePackage{fix-cm}
\documentclass{article}
\usepackage{spconf,amsmath,graphicx}
\usepackage{graphicx}
\usepackage[ruled,vlined]{algorithm2e}
\usepackage{amsmath}
\usepackage{amssymb}
\usepackage[colorlinks,linkcolor=blue]{hyperref}
\usepackage{algorithmic}
\usepackage{microtype}
\usepackage{graphicx}
\usepackage{subfigure}
\usepackage{booktabs} 
\usepackage{hyperref}

\usepackage{amsmath}
\usepackage{times}
\usepackage{graphicx}
\usepackage{color}
\usepackage{multirow}

\usepackage{mathtools} 
\usepackage{booktabs} 
\usepackage{color}
\usepackage{amsmath}
\usepackage{amssymb}
\usepackage{hyperref}
\usepackage{bm}

\usepackage{rotating}
\usepackage{bbm}
\usepackage{latexsym}
\usepackage{float}

\begin{document}

\title{Measuring the Transferability of $\ell_\infty$ Attacks by the $\ell_2$ Norm}
%
\name{$^{1}$Sizhe Chen, $^{2}$Qinghua Tao, $^{1}$Zhixing Ye, $^{1*}$Xiaolin Huang}
\address{$^{1}$Shanghai Jiao Tong University, $^{2}$KU Leuven.}
%
%
%
%
\maketitle
\begin{abstract}
Deep neural networks could be fooled by adversarial examples with trivial differences to original samples. To keep the difference imperceptible in human eyes, researchers bound the adversarial perturbations by the $\ell_\infty$ norm, which is now commonly served as the standard to align the strength of different attacks for a fair comparison. However, we propose that using the $\ell_\infty$ norm alone is not sufficient in measuring the attack strength, because even with a fixed $\ell_\infty$ distance, the $\ell_2$ distance also greatly affects the attack transferability between models. Through the discovery, we reach more in-depth understandings towards the attack mechanism, i.e., several existing methods attack black-box models better partly because they craft perturbations with 70\% to 130\% larger $\ell_2$ distances. Since larger perturbations naturally lead to better transferability, we thereby advocate that the strength of attacks should be simultaneously measured by both the $\ell_\infty$ and $\ell_2$ norm. Our proposal is firmly supported by extensive experiments on ImageNet dataset from 7 attacks, 4 white-box models, and 9 black-box models. 
\end{abstract}

\section{Introduction}
Deep learning has widely served as a mainstream tool in various fields, so its vulnerability has received great attention, among which Adversarial Examples (AEs) has been extensively studied \cite{goodfellow2014explaining, madry2017towards}. AEs are slightly perturbed from the original (clean) samples but cheat \emph{victim} Deep Neural Networks (DNNs) to produce incorrect predictions \cite{szegedy2013intriguing}. Also, white-box AEs on a \emph{surrogate} model can be transferred to cheat other unseen \emph{victim} models, i.e., the \emph{transfer-based attacks} \cite{dong2018boosting, xie2019improving}. 

In adversarial attacks, fair evaluations matter, and proper measurements of the attack strength are fundamental in distinguishing whether the performance improvements are brought by the settings (increasing the attack strength) or by the methods. The latter reveal underlying mechanisms of adversarial attacks, while the former simply bypasses the (inadequate) rules for good numerical results. By far, almost all transfer-based attacks bound the AEs by the $\ell_\infty$ ball with a fixed radius \cite{dong2018boosting, xie2019improving, dong2019evading, lin2019nesterov, chen2020universal}, based upon the assumption that the strength of attacks is maintained the same for a fair evaluation of the attack transferability on such occasions. However, an obvious counter-example could be: an attacker only crafts one single pixel \cite{su2019one}, while another attacker crafts all pixels to the $\ell_\infty$ bound \cite{goodfellow2014explaining}. Although their perturbations are the same under the $\ell_\infty$ measurement, their attack strength is apparently different.

In this paper, we propose that bounding the perturbations merely by the $\ell_\infty$ norm is insufficient, since other measurements, e.g., the $\ell_2$ norm, can also pose significant effects on the transferability of AEs with the same $\ell_\infty$-norm perturbations. Under the same $\ell_\infty$ budget, i.e., \emph{the $\ell_\infty$ attacks}, (1) attacks that produce AEs with larger $\ell_2$ distances from original samples result in higher-transferable AEs; (2) increasing the $\ell_2$ distances of AEs leads to a distinctive boost in transferability when the attack is samely conducted; (3) the transferability of AEs from almost all attacks stays amazingly the same when the $\ell_2$ distances are also fixed. These three phenomena firmly support our claim since they come from extremely large-scale experiments on ImageNet \cite{deng2009imagenet} from 7 attacks, 4 surrogate models, and 9 victim models with diverse architectures. 


Our proposal, although intuitive, takes the initial steps to introduce multiple norms to align attack strength for an unbiased evaluation and fair comparison, since the community has long neglected the $\ell_2$ distance for perturbations with a fixed $\ell_\infty$ norm. For example, Square attack \cite{andriushchenko2020square} and Fast Gradient Sign Method (FGSM) \cite{goodfellow2014explaining} attack black-box models efficiently \cite{su2018robustness} with 130\% larger $\ell_2$ distances compared to the popular Projected Gradient Descend (PGD) \cite{madry2017towards}. Similarly, variants of PGD \cite{dong2018boosting,lin2019nesterov} boost the transferability with 70\% larger $\ell_2$ distances, increasing the visibility under the same $\ell_\infty$ budget.

Since larger perturbations with greater visibility naturally lead to better transferability, we thereby advocate that the strength of all attacks should be measured by both the $\ell_\infty$ and $\ell_2$ norms. The reason and the benefit herein can be interpreted: the $\ell_\infty$ norm measures the maximal perturbation while the $\ell_2$ norm reflects the overall attack strength involving all pixels. The maximum and the ``average'' of perturbations both matter in measuring their saliency in human eyes.


\section{Methodology}
\subsection{Adversarial Attack and Its Transferability}\label{sec:advtrans}
In general, the adversarial attack can be described as
\begin{eqnarray}\label{adv}
\begin{split}
\left\{
\begin{array}{ll}
 {f(\boldsymbol x) \neq f(\boldsymbol x+\Delta \boldsymbol x)},  \\
{\|\Delta \boldsymbol x\|_{p} \leq \varepsilon},
\end{array}
\right.
\end{split}
\end{eqnarray}
where $f: \Omega \subset \mathbb R^n \mapsto y$ refers to the victim model which predicts differently on the original samples $\boldsymbol x_\mathrm{org} = \boldsymbol x$ and AEs $\boldsymbol x_\mathrm{adv} = \boldsymbol x + \Delta \boldsymbol x$. Here, $\Delta \boldsymbol x$ is the perturbation, which should be imperceptible in human eyes. In attack, we aim to fool DNNs by maximizing the training loss through optimizing the variable $\boldsymbol x$ with gradients $\boldsymbol g(\boldsymbol x)=\nabla_{\boldsymbol{x}} L(f(\boldsymbol x |\boldsymbol w), y)$. $\Delta \boldsymbol x$ is commonly restricted by $||\cdot||_p$, such as $\ell_1$-, $\ell_2$-, or $\ell_\infty$-norm, where $\ell_\infty$ is the most popular choice, and its iterative update of AE becomes $\boldsymbol x_{k+1} = \text{clip}^\varepsilon_{\boldsymbol x_\mathrm{org}}( \boldsymbol  x_{k} + \alpha \cdot \boldsymbol g_{k+1})$. 
When the complete knowledge (mostly the gradients) of victim DNNs is known, attackers could achieve high success rates \cite{goodfellow2014explaining, madry2017towards}.

However, in real-world scenarios, it is commonly impossible to attain the gradient information from the victim model, and thus black-box attacks become essential. It is discovered that the AEs from white-box attacks on a surrogate model could transfer to cheat other black-box victim models \cite{dong2018boosting, xie2019improving}. Various techniques have been proposed in enhancing attack transferability. The Momentum Iterative (MI) \cite{dong2018boosting} method first uses the momentum gradients, and later the Nesterov Iterative (NI) \cite{lin2019nesterov} chooses Nesterov accelerated gradients. In \cite{xie2019improving}, the Diverse Input is proposed to produce AEs using average gradients from randomly transformed input samples. The Translation Invariant \cite{dong2019evading} rather aims to generate AEs with respect to translated copies of the original samples. The Scale Invariant \cite{lin2019nesterov} updates AEs with average gradients from different scale copies of the input. It has also been proposed to attack the attention heatmap while maximizing the training loss \cite{chen2020universal}.



\subsection{Studied Transfer-Based Attacks}\label{sec:optat}
We investigate 7 transfer-based attacks, which have been verified to be effective concerning transferability. More importantly, they naturally lead to distinctively different $\ell_2$ distances by using different optimization algorithms (on AEs) so that the impacts of $\ell_2$ norms can be embodied clearly. 

Gradient Descent (\textbf{GD}) \cite{madry2017towards} performs the updates through the computed gradient direction. The corresponding attack is one-step FGSM \cite{goodfellow2014explaining} and multi-step PGD \cite{madry2017towards}. Momentum Gradient Descent (\textbf{M-GD}) \cite{dong2018boosting} accumulates the gradients in a moving-average way, so it implicitly optimizes with more one-stage information as in GD. As a result, it gains a faster convergence and reduces oscillation. Nesterov Accelerated Gradient Descent (\textbf{NA-GD}) \cite{lin2019nesterov} makes a big jump in the previously accumulated gradient and does a correction afterward. This predictive strategy prevents it from going too fast and leads to increased responsiveness. Besides storing a moving average of past squared gradients, \textbf{Adam} \cite{kingma2014adam} also keeps that of past gradients as in M-GD. The crucial idea is to go slower in dimensions that have gone far. \textbf{L-Adam} \cite{loshchilov2017decoupled} regularizes the gradients. The $\ell_2$ regularization is often referred to as the weight decay. Compared to Adam, \textbf{AdaBelief} \cite{zhuang2020adabelief} 
has better convergence on many tasks with no extra costs. Momentum Stochastic Variance-Adapted Gradient (\textbf{M-SVAG}) \cite{balles2018dissecting}, compared to M-GD, applies variance adaptation to update.

\subsection{Experimental Setups}\label{sec:setup}
In our studies, ImageNet \cite{deng2009imagenet} validation set with 50K images are used as original samples, following \cite{xie2019improving, lin2019nesterov, chen2020universal}. Keras pre-processing function, central cropping, and resizing (to 224 or 299) are adopted to pre-process the inputs. In all experiments, 1K images are randomly (with a fixed seed) selected and the images falsely predicted by the surrogate are skipped as in \cite{su2018robustness, chen2020universal}. Experiments are implemented in TensorFlow \cite{abadi2016tensorflow} with 4 NVIDIA GeForce RTX 2080Ti GPUs.

For attack and test, several well-trained models in Keras Applications are used. VGG16 \cite{simonyan2014very}, ResNet50 \cite{he2016deep}, InceptionV3 \cite{szegedy2016rethinking}, and DenseNet121 \cite{huang2017densely} are attacked as surrogate models. VGG19 \cite{simonyan2014very}, ResNet152 \cite{he2016deep}, DenseNet201 \cite{huang2017densely}, InceptionResNetV2 \cite{szegedy2017inception}, Xception \cite{chollet2017xception}, and NASNetLarge \cite{zoph2018learning} are chosen as black-box victims. To enrich the comparison, we also test on some publicly available adversarially-trained black box models, including InceptionV3adv \cite{kurakin2018adversarial}, InceptionResNetV2adv \cite{kurakin2018adversarial}, and ResNeXt101den \cite{xie2019feature}. The well-designed diversity of models can eliminate the bias from model architectures to a large extent, contributing to the solid conclusion concerning attacks and the $\ell_2$ distances.

The Attack Success Rate (ASR) is taken as the performance metric, which reflects the transferability and is measured by the average error rate of 9 victim models. The pre-average error rate and the code to reproduce our results are publicly available at \url{https://github.com/Sizhe-Chen/FairAttack}. Since the $\ell_2$ distance is dependent on the size of images, we use an associated normalized metric, which is the Root Mean Square Error (RMSE). RMSE calculates how far AEs perturb with respect to the $\ell_2$ distance, i.e., $dist(\boldsymbol x_\mathrm{adv}, \boldsymbol x_\mathrm{org}) = \sqrt{\|\boldsymbol x_\mathrm{adv}-\boldsymbol x_\mathrm{org}\|_2^{2} / N}$, where $N=H \cdot W \cdot C$ is the dimension of the sample.

To fairly evaluate the transferability of attacks, the optimization problem of the adversarial attack on the surrogate model is solved to a similar level with updates presented in Section \ref{sec:advtrans}. That is to say, the loss of the surrogate on AEs is controlled to a fixed value (such as 0.03 in our experiments if not otherwise stated). All perturbations are $\ell_\infty$-bounded by $\varepsilon=16$, following \cite{xie2019improving, dong2019evading, lin2019nesterov}.

\section{$\ell_2$ Distance Affects $\ell_\infty$ Attacks}\label{sec:affect} 
In this section, we reveal the current insufficient measurement of the attack strength by showing the influence of the $\ell_2$ norm on the transferability via experiments on 7 transfer-based attacks, 4 surrogates, and 9 victims. Under the same $\ell_\infty$ bound, (1) attacks that optimize AEs to larger $\ell_2$ distances enjoy better transferability; (2) enlarging $\ell_2$ distances of AEs leads to a steady increase of transferability; (3) fixing the $\ell_2$ distances of AEs produces amazingly consistent transferability for all attacks. Observations above solidly validate the deficiency of current attack strength measurements by solely the $\ell_\infty$ norm.

\subsection{Transferability Correlates with RMSE}\label{sec:corr}
\begin{figure}
  \centering
  \includegraphics[width=\linewidth]{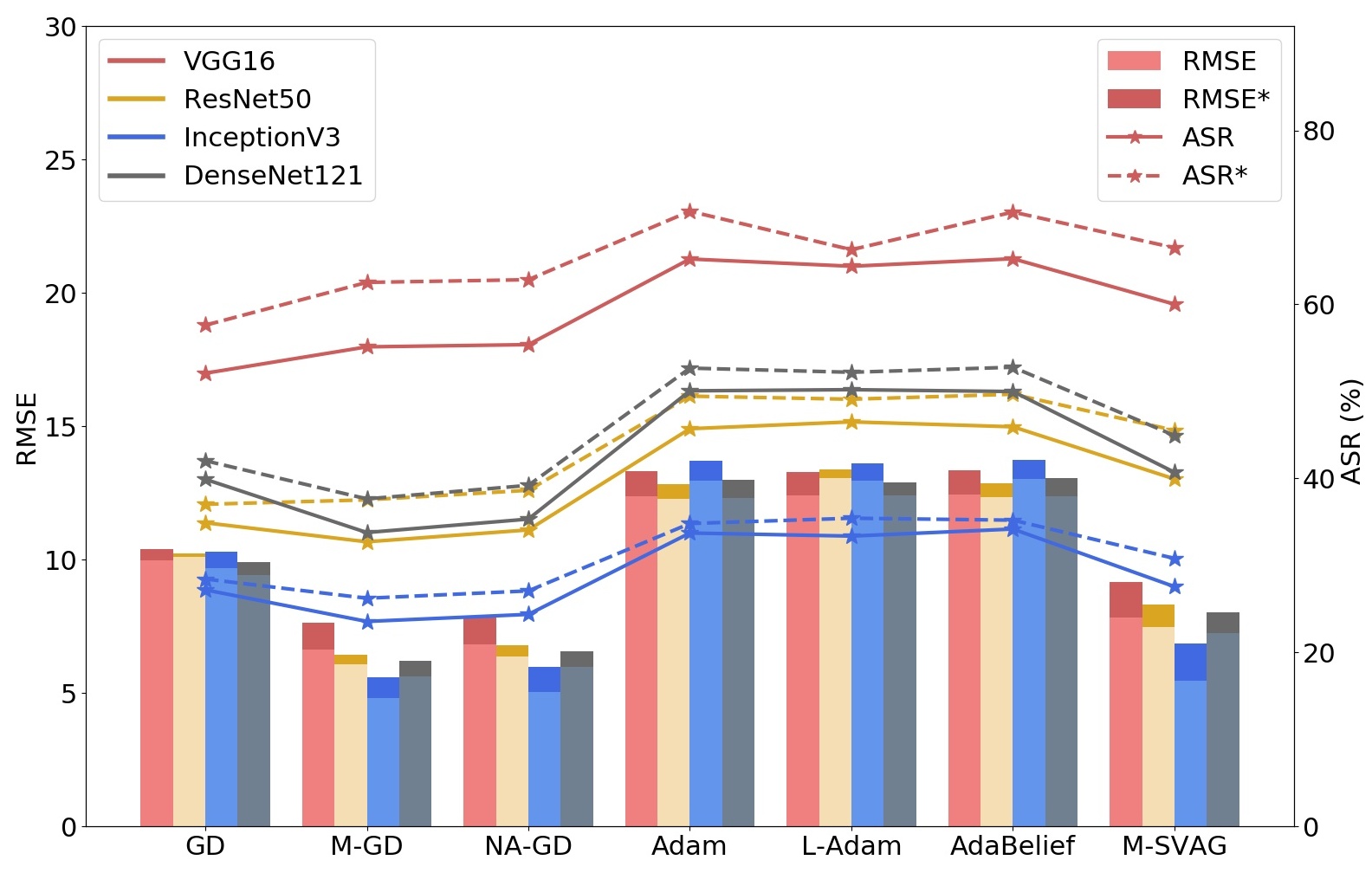}
  \caption{Transferability (lines) and RMSE (bars) of the AEs crafted by 7 attacks. 
  Solid lines and light bars are results for Sec. \ref{sec:corr}. Dotted lines (ASR*) and the shadowed dark bars (RMSE*, on top of the bars) are results with the additional loss (\ref{eq:loss:added}) for Sec. \ref{sec:rmseloss}, where the AEs reach a larger RMSE when the attack stops at the same cross-entropy loss.
  }
  \label{fig:exp123}
\end{figure}

We first empirically study the transferability under the settings in Sec. \ref{sec:setup}, and Fig. \ref{fig:exp123} plots the results. Two indices are worth mentioning and comparing, i.e., the ASR (transferability) and the RMSE of AEs crafted by the tested 7 attacks. 

For the solid lines (for transferability) and their corresponding light bars (for RMSE) in the same color, one could observe that the transferability has a strong correlation with RMSE. For the transferability, Adam family outperforms others, and then M-SVAG follows up. More specifically, the resulting transferability is ranked: Adam family $\textgreater$ M-SVAG $\textgreater$ GD $\textgreater$ NA-GD $\approx$ M-GD. For RMSE, the Adam family also stands out, and a similar rank is obtained: Adam family $\textgreater$ GD $\textgreater$ M-SVAG $\textgreater$ NA-GD $\approx$ M-GD. Among those surveyed attacks, except a slight difference concerning GD, this phenomenon holds true for all tested attacks. More notably, this overall trend remains stable for all 4 surrogates. 

The results show that when the attack is equally conducted on surrogates, the ASR (transferability) of a certain attack has a strong correlation with the corresponding $\ell_2$ distances between AEs and their original samples, indicating that RMSE is an important factor for the transferability. To further validate this hypothesis, we consider directly intervening in the RMSE and investigating its effect on transferability.

\begin{figure}[ht!]
  \centering
  \includegraphics[width=\linewidth]{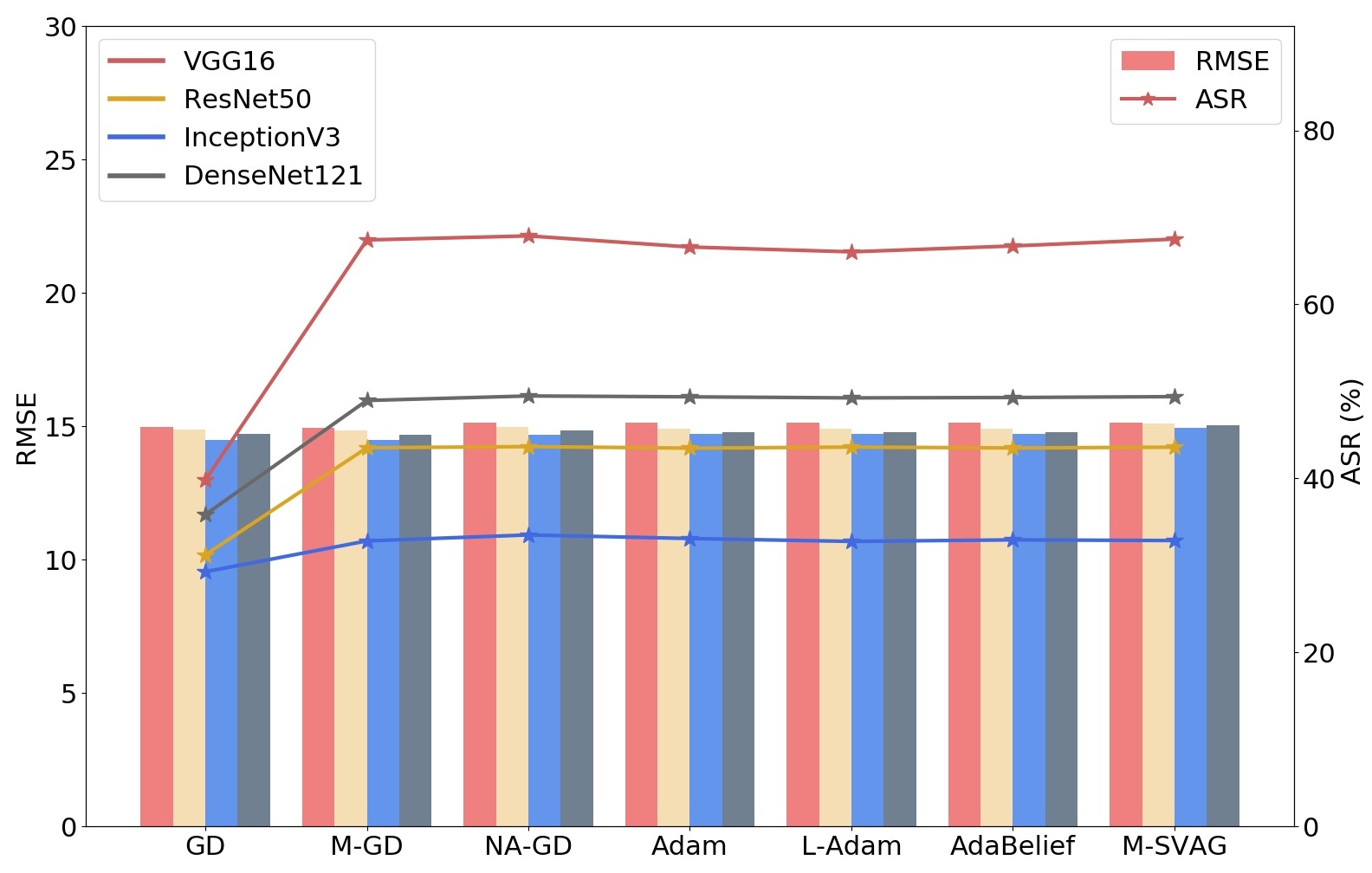}
  \caption{Transferability (lines) and RMSE (bars) of the AEs crafted by 7 attacks with a constant RMSE, i.e., $dist(\boldsymbol x, \boldsymbol x_\mathrm{org})=15$.}
  \label{fig:exp3.5}
\end{figure}

\subsection{Transferability with Larger RMSE}\label{sec:rmseloss}
If the hypothesis remains true,  the transferability should be boosted by straightforwardly increasing the $\ell_2$ distance (RMSE) with other settings fixed. In fact, larger RMSE can be gained in different ways, and here we offer one alternative by supplementing an RMSE term to the original loss, i.e., 
\begin{eqnarray}\label{eq:loss:added}
L'(f( \boldsymbol x| \boldsymbol{w}), y)=L(f(\boldsymbol x| \boldsymbol{w}), y)+ \gamma \|\boldsymbol x - \boldsymbol x_\mathrm{org}\|_2,
\end{eqnarray}
where $L(f(\boldsymbol x| \boldsymbol{w}), y)$ is the standard cross-entropy loss, $\|\boldsymbol x - \boldsymbol x_\mathrm{org}\|_2$ is the added RMSE between the potential AEs and their original samples, and the parameter $\gamma > 0$ controls the trade-off between the two parts in the loss.

With the modified loss function (\ref{eq:loss:added}), the influence of RMSE term $\|\boldsymbol x - \boldsymbol x_\mathrm{org}\|_2$ can now be evaluated in a quantitative way, maintaining the cross-entropy $L(f(\boldsymbol x| \boldsymbol{w}), y)$ to a fixed value. Under this setting, we extend the experiment of Sec. \ref{sec:affect} by performing the attacks with the modified loss (\ref{eq:loss:added}) until the cross-entropy $L(f(\boldsymbol x| \boldsymbol{w}), y)$ reaches the same level, such that the RMSE becomes the only factor that varies. We can see from Fig. \ref{fig:exp123} that a definite boost in transferability (from the solid lines to the dotted lines) is obtained under all circumstances, which strongly supports our hypothesis. 


\subsection{Transferability with Fixed RMSE}\label{sec:fixrmse}
Besides the additional term to induce a large RMSE, another possible validation is to directly control the RMSE to the same level, which should also produce a similar ASR if our hypothesis is correct, i.e., RMSE is indeed an important metric in evaluating the attack strength in $\ell_\infty$ attacks.

Therefore, we carefully tune the hyper-parameters to maintain AEs to have a similar RMSE $dist(\boldsymbol x, \boldsymbol x_\mathrm{org})$ by stopping the attack when AEs approach a pre-set RMSE of 15 and Fig. \ref{fig:exp3.5} reports the results. Surprisingly, the transferability reaches a similar value for all surrogates with almost all attacks. The 4 flat lines indicate that the ASR towards 9 victims is found alike, however the specific optimization proceeds. Because the RMSE is imposed to be fixed, different optimization algorithms hardly impact the transferability, which is revealed and demonstrated by almost all studies in this paper.

This encouraging discovery also positively supports our hypothesis that attackers could impact the transferability via RMSE because if the RMSE is kept fixed, various attacks could not the transferability, i.e., the powerless curves failing to oscillate in Fig. \ref{fig:exp3.5}.

\begin{figure}
  \centering
  \includegraphics[width=\linewidth]{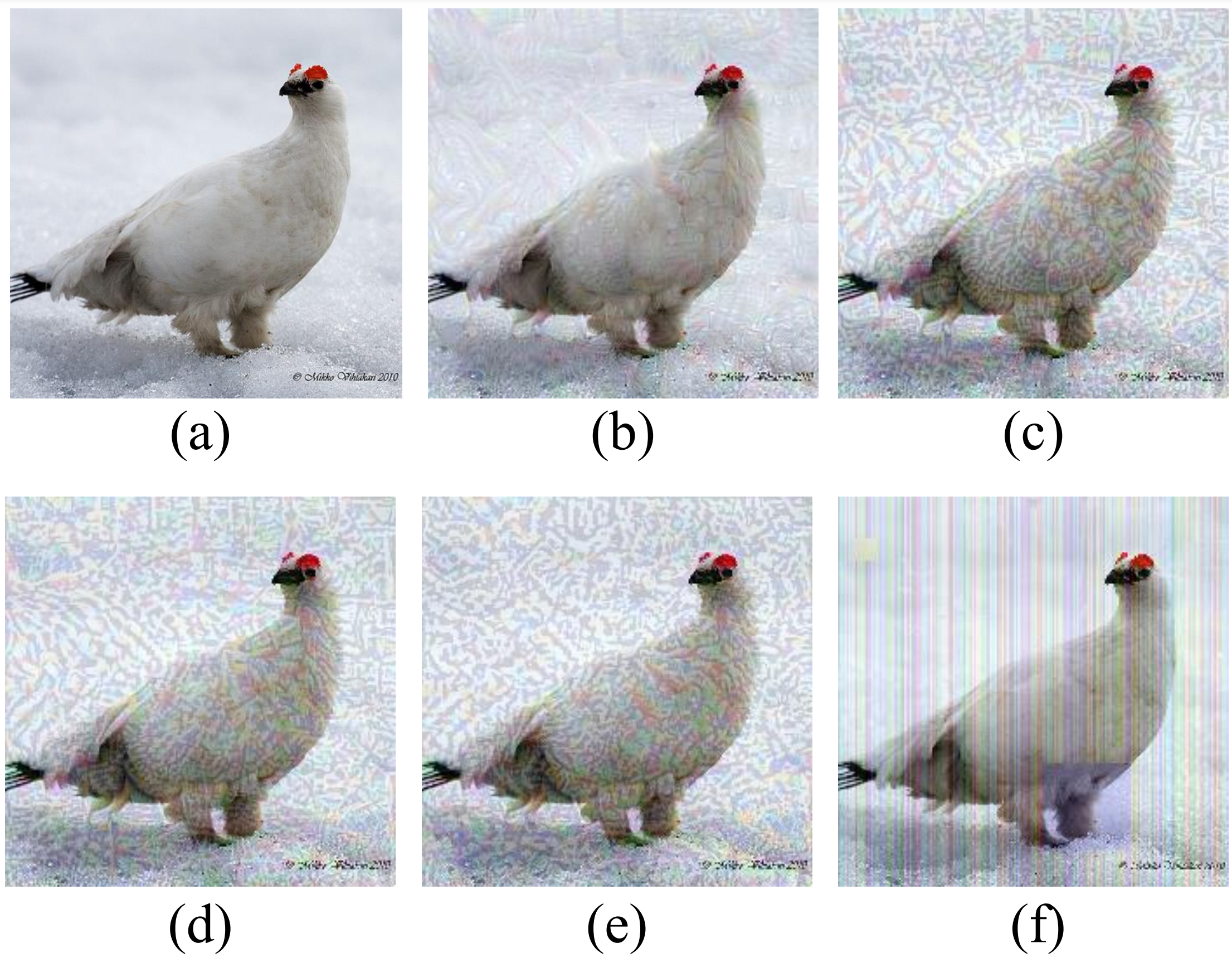}
  \caption{AEs from small-perturbed (b) and large-perturbed attacks (c), (d), (e), (f). (a) is the original image; (b) is the PGD AE; (c) is the MI-PGD AE; (d) is the NI-PGD AE; (e) is the FGSM AE; (f) is the Square attack AE. The AEs are crafted by attacking VGG16 \cite{simonyan2014very} within $\ell_\infty$ bound 16.}
  \label{fig:exp5}
\end{figure}

\section{Discussions on Fair Evaluations}
In the aforementioned extensive experiments, we have empirically verified that the sole $\ell_\infty$-norm is insufficient for measuring the attack strength since $\ell_2$ distances (measured by RMSE) also greatly influence the attack performance. In this section, we discuss why and how this seemingly intuitive conclusion is long-neglected in the community, and how we could perform a more unbiased evaluation on adversarial attacks.


The reason why the sole $\ell_\infty$ measurement is inadequate is that iterative gradient-based attacks, such as PGD \cite{madry2017towards}, generally yield lots of insufficiently-perturbed pixels because the attack gradients are mostly sparse \cite{chen2020universal}, i.e., extremely large in few pixels and small in most pixels. For example, PGD generally produces AEs with RMSE around 7 under the $\ell_\infty$ bound $\varepsilon=16$, which is far from the maximum RMSE distance of 16 if all pixels are fully perturbed. Since $\ell_\infty$-norm focuses only on the maximum, using this measurement prevents us from discovering the insufficiently-perturbed pixels. In contrast, the $\ell_2$ distance considers all pixels, so we could distinguish different attack strengths more clearly.




Because there exist many insufficiently perturbed pixels in PGD, methods that keep perturbing those pixels tend to boost the transferability as shown in Sec. \ref{sec:rmseloss}, which is however not preferred due to the greater visibility. For example, MI-PGD \cite{dong2018boosting} and NI-PGD \cite{lin2019nesterov} mostly produce AEs with RMSE around 12, which is significantly larger than that in PGD, making the AEs more perceivable. We could also explain why FGSM \cite{goodfellow2014explaining} transfers better than PGD \cite{madry2017towards} as shown in \cite{su2018robustness}, a contradictory phenomenon since PGD is supposed to transfer better for its greater performance in white-box settings with more diverse patterns \cite{chen2020universal}. The reason here is that FGSM perturbs pixels only once, so it extremely exploits the feasible values within the $\ell_\infty$ bound to push every pixel to the limit, leading to the maximal RMSE. Another example is the Square attack \cite{andriushchenko2020square}, which generates the maximally perturbed (under the $\ell_\infty$ bound) AEs for each query, so its great performance may be partly due to this exhausted perturbing strategy. We visualize AEs from small perturbed attacks and large perturbed ones in Fig. \ref{fig:exp5}, where the attack strength and the visibility of AEs differ a lot even they are bounded under the same $\ell_\infty$ norm.


In most of the existing attack experiments, e.g., \cite{dong2018boosting, xie2019improving, dong2019evading, lin2019nesterov}, the attack strength is only measured by the $\ell_\infty$ norm. However, as verified in the above experiments, one could change the $\ell_2$ distance to get different attack performance within the same $\ell_\infty$ bound. In this concern, we advocate measuring the strength of all attacks by both the $\ell_\infty$ and $\ell_2$ distance. In other words, to fairly evaluate the attack transferability, one could at least maintain the same $\ell_2$ distance under a fixed $\ell_\infty$ bound in $\ell_\infty$ attack and also report the RMSE as done in \cite{chen2020universal}.


\section{Conclusion}
In this paper, we aim at investigating the informative measurement of the attack strength when evaluating the transferability, and find out that the existing practice to only bound the $\ell_\infty$ norm is insufficient. 
To support this, we extensively explore the influence of the $\ell_2$ distance of AEs on the attack transferability, which is however long-neglected in $\ell_\infty$ attacks despite the intuitiveness. 
Accordingly, attackers are supposed to maintain the $\ell_2$ distance in attacks even the $\ell_\infty$ distance is already fixed. Otherwise, attackers may trickily bypass the measurement to seemingly enhance the transferability by actually making the AEs more perceivable. Our work takes the first step giving rise to the attention for fair evaluations on the attack strength, and the $\ell_2$ norm is just one metric aside of the $\ell_\infty$ norm and other metrics can be also influential, e.g., the $\ell_0$ norm, the $\ell_1$ norm, SSIM, the distance in feature spaces.

\section*{Acknowledgements}
This work is partly supported by the National Natural Science Foundation of China (61977046), Shanghai Science and Technology Program (22511105600), and Shanghai Municipal Science and Technology Major Project (2021SHZDZX0102).

\bibliographystyle{IEEEtran} 
\bibliography{IEEEabrv, Reference}

\begin{thebibliography}{10}
\providecommand{\url}[1]{#1}
\csname url@samestyle\endcsname
\providecommand{\newblock}{\relax}
\providecommand{\bibinfo}[2]{#2}
\providecommand{\BIBentrySTDinterwordspacing}{\spaceskip=0pt\relax}
\providecommand{\BIBentryALTinterwordstretchfactor}{4}
\providecommand{\BIBentryALTinterwordspacing}{\spaceskip=\fontdimen2\font plus
\BIBentryALTinterwordstretchfactor\fontdimen3\font minus
  \fontdimen4\font\relax}
\providecommand{\BIBforeignlanguage}[2]{{%
\expandafter\ifx\csname l@#1\endcsname\relax
\typeout{** WARNING: IEEEtran.bst: No hyphenation pattern has been}%
\typeout{** loaded for the language `#1'. Using the pattern for}%
\typeout{** the default language instead.}%
\else
\language=\csname l@#1\endcsname
\fi
#2}}
\providecommand{\BIBdecl}{\relax}
\BIBdecl

\bibitem{goodfellow2014explaining}
I.~J. Goodfellow, J.~Shlens, and C.~Szegedy, ``Explaining and harnessing
  adversarial examples,'' in \emph{Int. Conf. on Learn. Rep.}, 2015.

\bibitem{madry2017towards}
A.~Madry, A.~Makelov, L.~Schmidt, D.~Tsipras, and A.~Vladu, ``Towards deep
  learning models resistant to adversarial attacks,'' in \emph{Int. Conf. on
  Learn. Rep.}, 2018.

\bibitem{szegedy2013intriguing}
C.~Szegedy, W.~Zaremba, I.~Sutskever, J.~Bruna, D.~Erhan, I.~J. Goodfellow, and
  R.~Fergus, ``Intriguing properties of neural networks,'' in \emph{Int. Conf.
  on Learn. Rep.}, 2014.

\bibitem{dong2018boosting}
Y.~Dong, F.~Liao, T.~Pang, H.~Su, J.~Zhu, X.~Hu, and J.~Li, ``Boosting
  adversarial attacks with momentum,'' in \emph{IEEE Conf. on Comp. Vis. and
  Patt. Recog.}, 2018, pp. 9185--9193.

\bibitem{xie2019improving}
C.~Xie, Z.~Zhang, Y.~Zhou, S.~Bai, J.~Wang, Z.~Ren, and A.~L. Yuille,
  ``Improving transferability of adversarial examples with input diversity,''
  in \emph{IEEE Conf. on Comp. Vis. and Patt. Recog.}, 2019, pp. 2730--2739.

\bibitem{dong2019evading}
Y.~Dong, T.~Pang, H.~Su, and J.~Zhu, ``Evading defenses to transferable
  adversarial examples by translation-invariant attacks,'' in \emph{IEEE Conf.
  on Comp. Vis. and Patt. Recog.}, 2019, pp. 4312--4321.

\bibitem{lin2019nesterov}
J.~Lin, C.~Song, K.~He, L.~Wang, and J.~E. Hopcroft, ``Nesterov accelerated
  gradient and scale invariance for adversarial attacks,'' in \emph{Int. Conf.
  on Learn. Rep.}, 2020.

\bibitem{chen2020universal}
S.~Chen, Z.~He, C.~Sun, and X.~Huang, ``Universal adversarial attack on
  attention and the resulting dataset {DAmageNet},'' in \emph{IEEE Trans. Patt.
  Anal. and Mach. Intell.}, 2020, pp. 1--1.

\bibitem{su2019one}
J.~Su, D.~V. Vargas, and K.~Sakurai, ``One pixel attack for fooling deep neural
  networks,'' in \emph{IEEE Trans. on Evo. Comput.}, 2019, pp. 828--841.

\bibitem{deng2009imagenet}
J.~Deng, W.~Dong, R.~Socher, L.-J. Li, K.~Li, and L.~Fei-Fei, ``Imagenet: A
  large-scale hierarchical image database,'' in \emph{IEEE Conf. on Comp. Vis.
  and Patt. Recog.}, 2009, pp. 248--255.

\bibitem{andriushchenko2020square}
M.~Andriushchenko, F.~Croce, N.~Flammarion, and M.~Hein, ``Square attack: A
  query-efficient black-box adversarial attack via random search,'' in
  \emph{Eur. Conf. on Comp. Vis.}, 2020, pp. 484--501.

\bibitem{su2018robustness}
D.~Su, H.~Zhang, H.~Chen, J.~Yi, P.-Y. Chen, and Y.~Gao, ``Is robustness the
  cost of accuracy?--a comprehensive study on the robustness of 18 deep image
  classification models,'' in \emph{Eur. Conf. on Comp. Vis.}, 2018, pp.
  631--648.

\bibitem{kingma2014adam}
D.~P. Kingma and J.~Ba, ``Adam: {A} method for stochastic optimization,'' in
  \emph{Int. Conf. on Learn. Rep.}, 2015.

\bibitem{loshchilov2017decoupled}
I.~Loshchilov and F.~Hutter, ``Decoupled weight decay regularization,''
  \emph{arXiv preprint arXiv:1711.05101}, 2017.

\bibitem{zhuang2020adabelief}
J.~Zhuang, T.~Tang, S.~Tatikonda, N.~Dvornek, Y.~Ding, X.~Papademetris, and
  J.~S. Duncan, ``Adabelief optimizer: Adapting stepsizes by the belief in
  observed gradients,'' in \emph{Adv. in Neural Info. Process. Sys.}, 2020, pp.
  18\,795--18\,806.

\bibitem{balles2018dissecting}
L.~Balles and P.~Hennig, ``Dissecting adam: The sign, magnitude and variance of
  stochastic gradients,'' in \emph{Int. Conf, on Mach, Learn.}, 2018, pp.
  404--413.

\bibitem{abadi2016tensorflow}
M.~Abadi, A.~Agarwal, P.~Barham, E.~Brevdo, Z.~Chen, C.~Citro, G.~S. Corrado,
  A.~Davis, J.~Dean, M.~Devin \emph{et~al.}, ``{Tensorflow}: Large-scale
  machine learning on heterogeneous distributed systems,'' \emph{arXiv preprint
  arXiv:1603.04467}, 2016.

\bibitem{simonyan2014very}
K.~Simonyan and A.~Zisserman, ``Very deep convolutional networks for
  large-scale image recognition,'' in \emph{Int. Conf. on Learn. Rep.}, 2015.

\bibitem{he2016deep}
K.~He, X.~Zhang, S.~Ren, and J.~Sun, ``Deep residual learning for image
  recognition,'' in \emph{IEEE Conf. on Comp. Vis. and Patt. Recog.}, 2016, pp.
  770--778.

\bibitem{szegedy2016rethinking}
C.~Szegedy, V.~Vanhoucke, S.~Ioffe, J.~Shlens, and Z.~Wojna, ``Rethinking the
  inception architecture for computer vision,'' in \emph{IEEE Conf. on Comp.
  Vis. and Patt. Recog.}, 2016, pp. 2818--2826.

\bibitem{huang2017densely}
G.~Huang, Z.~Liu, L.~van~der Maaten, and K.~Q. Weinberger, ``Densely connected
  convolutional networks,'' in \emph{IEEE Conf. on Comp. Vis. and Patt.
  Recog.}, 2017, pp. 2261--2269.

\bibitem{szegedy2017inception}
C.~Szegedy, S.~Ioffe, V.~Vanhoucke, and A.~A. Alemi, ``Inception-v4,
  inception-resnet and the impact of residual connections on learning,'' in
  \emph{AAAI Conf. on Artif. Intell.}, 2017, pp. 4278--4284.

\bibitem{chollet2017xception}
F.~Chollet, ``Xception: Deep learning with depthwise separable convolutions,''
  in \emph{IEEE Conf. on Comp. Vis. and Patt. Recog.}, 2017, pp. 1251--1258.

\bibitem{zoph2018learning}
B.~Zoph, V.~Vasudevan, J.~Shlens, and Q.~V. Le, ``Learning transferable
  architectures for scalable image recognition,'' in \emph{IEEE Conf. on Comp.
  Vis. and Patt. Recog.}, 2018, pp. 8697--8710.

\bibitem{kurakin2018adversarial}
A.~Kurakin, I.~Goodfellow, S.~Bengio, Y.~Dong, F.~Liao, M.~Liang, T.~Pang,
  J.~Zhu, X.~Hu, C.~Xie \emph{et~al.}, ``Adversarial attacks and defences
  competition,'' in \emph{The NIPS'17 Competition: Building Intelligent
  Systems}, 2018.

\bibitem{xie2019feature}
C.~Xie, Y.~Wu, L.~v.~d. Maaten, A.~L. Yuille, and K.~He, ``Feature denoising
  for improving adversarial robustness,'' in \emph{IEEE Conf. on Comp. Vis. and
  Patt. Recog.}, 2019, pp. 501--509.

\end{thebibliography}
\end{document}